\documentclass[letterpaper, 10 pt, conference]{ieeeconf}  

\IEEEoverridecommandlockouts                              

\usepackage{graphics} 
\usepackage{epsfig} 
\usepackage{times} 
\usepackage{amsmath} 
\usepackage{amssymb}  
\usepackage{amsfonts}
\usepackage[colorlinks,linkcolor=black,anchorcolor=black,citecolor=black,urlcolor=black]{hyperref}
\usepackage[table]{xcolor}
\usepackage{booktabs}
\usepackage{multirow}
\usepackage{graphicx}
\usepackage{subfigure}
\usepackage{cite}
\usepackage{float}
\usepackage{makecell}
\usepackage{algorithm}
\usepackage{algpseudocode}
\usepackage{etoolbox}
\usepackage{graphicx}
\usepackage{float} 
\usepackage{subfigure}
\usepackage{titlesec}
\usepackage{cite}

\titleformat{\subsubsection}[runin]
{\normalfont\bfseries} 
{}
{0em} 
{}

\titlespacing*{\subsubsection}
{0pt}{3.25ex plus 1ex minus .2ex}{0.5em plus .2ex}
\titlespacing*{\subsubsection}{0pt}{\baselineskip}{0.5\baselineskip}

\makeatletter
\patchcmd{\@makecaption}
  {\scshape}
  {}
  {}
  {}
\makeatletter
\patchcmd{\@makecaption}
  {\\}
  {.\ }
  {}
  {}
\makeatother

\graphicspath{{imgs/}}

\title{\LARGE \bf
OnFly: Onboard Zero-Shot Aerial Vision-Language Navigation \\ toward Safety and Efficiency
}

\author{
Guiyong Zheng$^{1,2}$,
Yueting Ban$^{2}$,
Mingjie Zhang$^{3,2}$,
Juepeng Zheng$^{1}$,
and Boyu Zhou$^{2,\dag}$
\thanks{\textbf{$^{\dag}$ Corresponding Author}}
\thanks{$^{1}$School of Artificial Intelligence, Sun Yat-Sen University, Zhuhai, China.}
\thanks{$^{2}$Southern University of Science and Technology, Shenzhen, China}
\thanks{$^{3}$The Hong Kong University of Science and Technology, Guangzhou, China.}
\thanks{Email:}
\thanks{ {\tt\footnotesize zhouby@sustech.edu.cn}} 
}

\providecommand{\keywords}[1]{\textbf{\textit{Index terms---}} #1}
\begin{document}
\maketitle


\begin{abstract}

Aerial vision-language navigation (AVLN) enables UAVs to follow natural-language instructions in complex 3D environments. However, existing zero-shot AVLN methods often suffer from unstable single-stream Vision-Language Model decision-making, unreliable long-horizon progress monitoring, and a trade-off between safety and efficiency. We propose OnFly, a fully onboard, real-time framework for zero-shot AVLN. 
OnFly adopts a shared-perception dual-agent architecture that decouples high-frequency target generation from low-frequency progress monitoring, thereby stabilizing decision-making. It further employs a hybrid keyframe-recent-frame memory to preserve global trajectory context while maintaining KV-cache prefix stability, enabling reliable long-horizon monitoring with termination and recovery signals. In addition, a semantic-geometric verifier refines VLM-predicted targets for instruction consistency and geometric safety using VLM features and depth cues, while a receding-horizon planner generates optimized collision-free trajectories under geometric safety constraints, improving both safety and efficiency.
In simulation, OnFly improves task success from 26.4\% to 67.8\%, compared with the strongest state-of-the-art baseline, while fully onboard real-world flights validate its feasibility for real-time deployment. The code will be released at \url{https://github.com/Robotics-STAR-Lab/OnFly}
\end{abstract}

\section{Introduction}
\label{sec:intro}

Aerial vision-language navigation (AVLN) empowers UAVs to navigate complex 3D scenes following free-form natural-language instructions \cite{hu2025see,nasiriany2024pivot,gao2025openfly,liu2023aerialvln,sautenkov2025uav}. As a pivotal paradigm linking human intent understanding and autonomous UAV flight, AVLN has garnered extensive attention in academia and industry, with broad application prospects in intelligent inspection, emergency response and smart city management. Recently, powered by the strong generalization of vision-language models (VLMs), a series of zero-shot AVLN frameworks have emerged, greatly expanding the applicable scenarios of AVLN systems in unseen open-world environments \cite{hu2025see,chen2023typefly,lykov2025cognitivedrone,zhang2025logisticsvln,nasiriany2024pivot}.

Despite promising advances, zero-shot AVLN methods face several challenges impeding real-world deployment. First, the single-stream VLM decision paradigm suffers from critical disparate-frequency demand coupling and multi-objective misalignment, leading to inherent decision instability. Most approaches couple high-frequency real-time flight and low-frequency task progress monitoring into one inference stream, which causes mutual interference, slows target output, and degrades overall performance \cite{hu2025see,zhang2025logisticsvln,lykov2025cognitivedrone}. Second, traditional temporal memory designs for progress monitoring have critical flaws: sliding-window memory gradually loses global trajectory context as new observations are added, and frequent overwriting of early slots breaks KV-cache prefix stability \cite{shi2024keep}, leading to low cache reuse rates, higher inference latency, and unreliable monitoring that triggers task failures. Lastly, flight safety and navigation efficiency present an unresolved contradiction. To mitigate risks from observation noise and environmental uncertainty, existing methods rely on conservative short-horizon actions via iterative inference, which avoids collisions but incurs severe stop-and-go behavior from long inference latency, drastically reducing efficiency \cite{nasiriany2024pivot,chen2023typefly}. Conversely, aggressive long-range target prediction boosts efficiency by extending single-step flight scope \cite{hu2025see}. However, the lack of systematic safety mechanisms makes UAVs prone to collisions.

\begin{figure}[t]
    \vspace{0.2cm}
        \begin{center}
        \includegraphics[width=0.99\columnwidth]{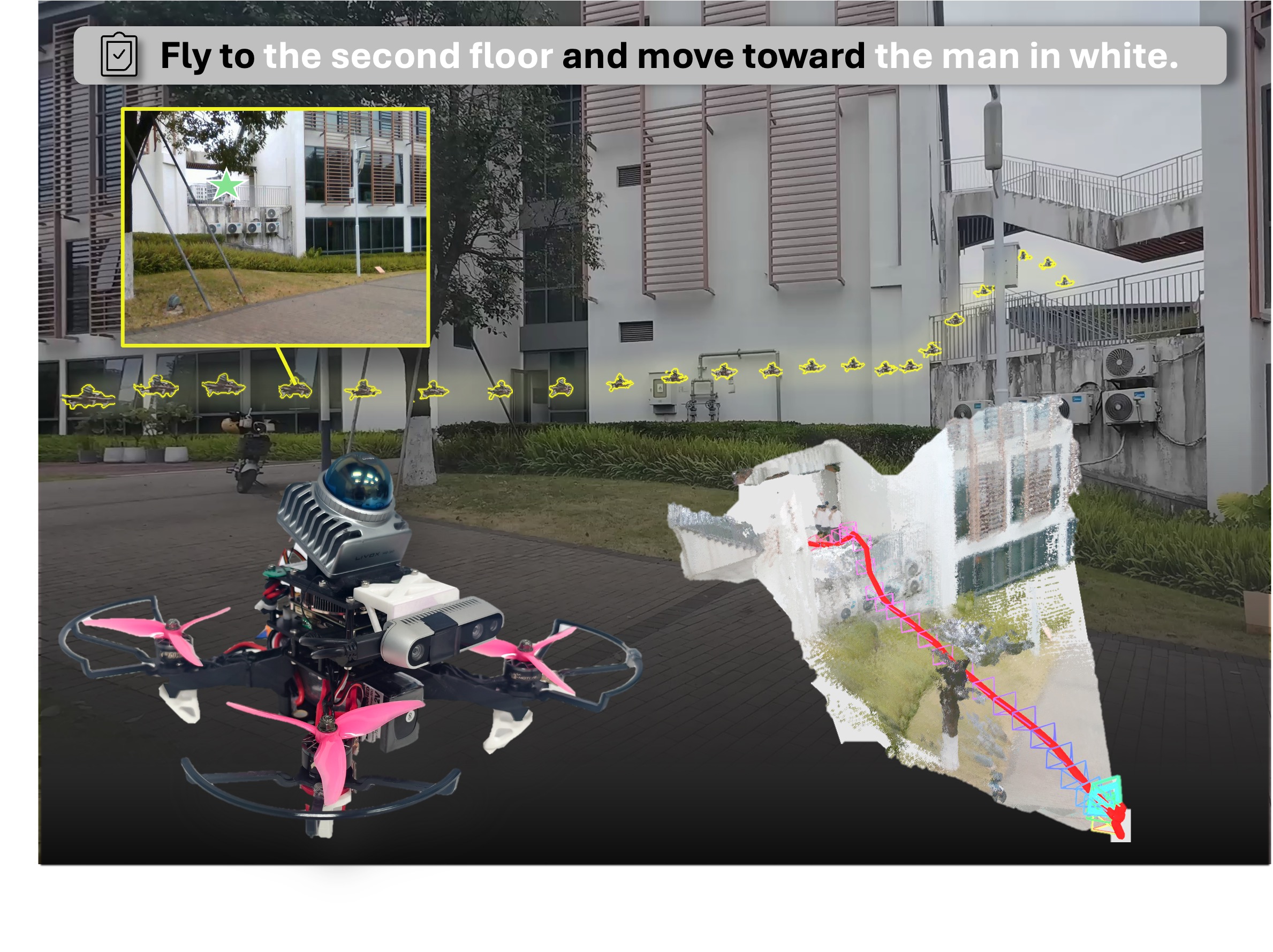}     
        \vspace{-1.5cm}
	   \end{center}
        \caption{\label{fig:top}\textbf{Real-world example of onboard language-guided UAV navigation.} The figure illustrates the UAV platform (left), the first-person onboard view (inset), and the executed flight trajectory (right).}
   \vspace{-0.7cm}
\end{figure}

\begin{figure*}[t]
    \vspace{0.cm}
        \begin{center}
        \includegraphics[width=0.99\textwidth]{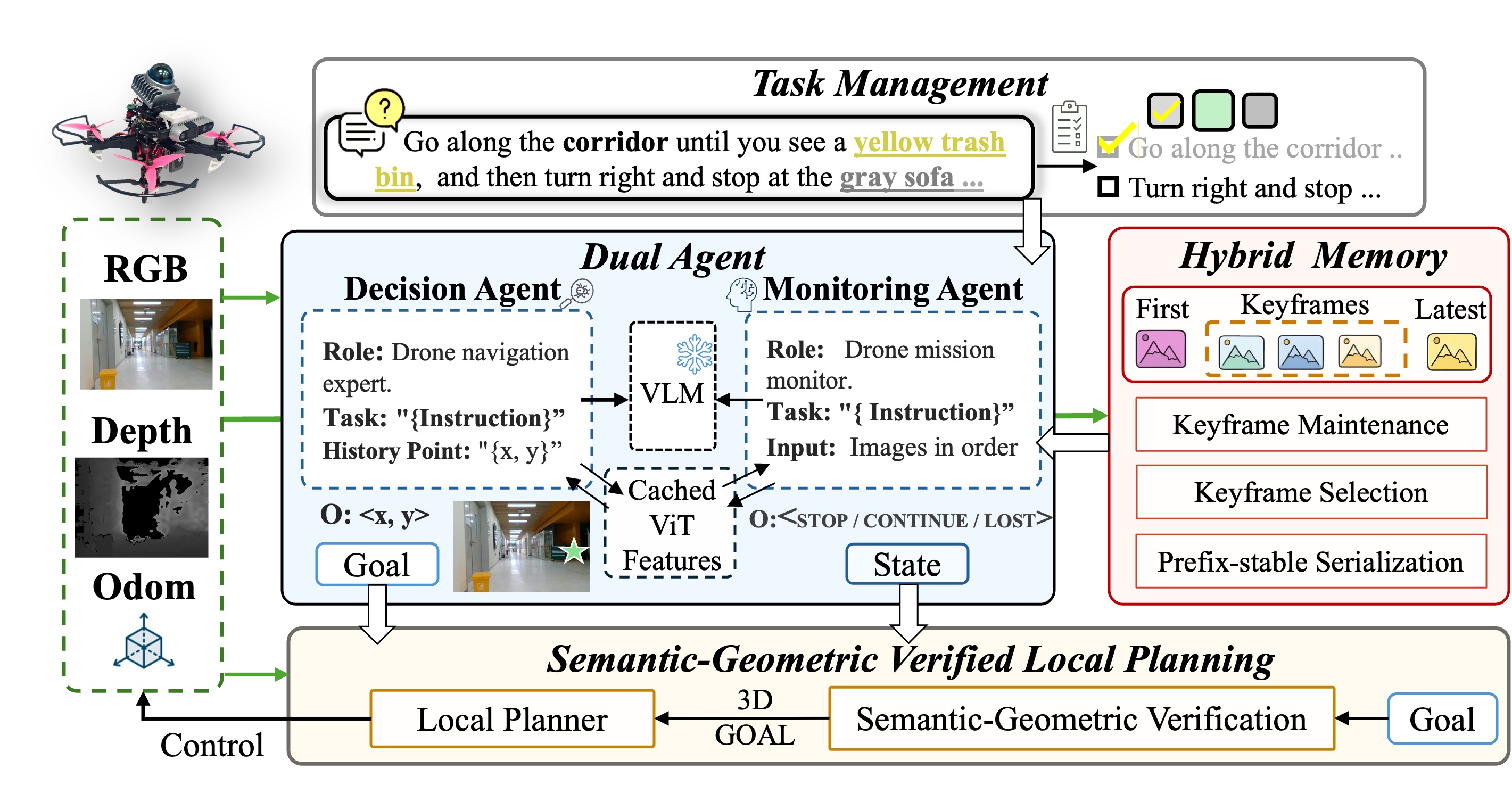}     
        \vspace{-0.6cm}
	   \end{center}
        \caption{\label{fig:system}\textbf{System pipeline of OnFly.} OnFly is a fully onboard system for AVLN with dual-agent decision-making and hybrid memory-based progress monitoring. The task manager splits a long instruction into subtasks and outputs the current one. The dual-agent module predicts candidate goals and monitors progress from onboard observations, where the \textit{Hybrid Memo}ry consists of the initial frame, keyframes, and the latest frame, to preserve global trajectory context and KV-cache prefix stability. Semantic--geometric verification and a safety-aware planner then generate an executable real-time trajectory.}
    \vspace{-0.5cm}
\end{figure*}

To tackle these intertwined challenges, we present \textbf{OnFly}, a fully onboard zero-shot real-time AVLN system with three targeted innovative designs that align with the above bottlenecks. First, we propose a shared-perception dual-agent architecture to decouple conflicting demands: the two agents share a Vision Transformer (ViT) backbone and cached visual features to eliminate redundant computation, and maintain independent KV-caches for their respective update rates, with a high-frequency decision agent generating real-time flight targets for responsiveness and a low-frequency monitoring agent tracking progress independently, avoiding inference interference and stabilizing overall decision-making. Second, we design a custom hybrid keyframe-recent-frame memory for the monitoring agent, which constructs memory with the initial frame, representative keyframes, and the latest real-time frame. It preserves global trajectory context and ensures KV-cache prefix stability via keyframe deduplication and serialization, enabling reliable long-horizon monitoring and accurate output of critical termination/recovery signals. Third, we integrate a semantic-geometric verifier with a receding-horizon local planner. The verifier validates and refines targets for instruction consistency and geometric safety via VLM feature matching and depth constraints. The planner leverages an ESDF map for obstacle avoidance and yaw regularization to generate optimized collision-free trajectories for verified long-range targets, achieving a robust balance between flight safety and navigation efficiency.

We evaluate OnFly in simulation and fully onboard real-world flight scenarios. In simulation, OnFly improves task success \textbf{from 26.4\% to 67.8\%} over the strongest baseline, and real-world onboard experiments further validate its efficiency and flight safety. We will release our implementation to facilitate future research on deployable AVLN. Our contributions are summarized as follows:

\begin{enumerate}
    \item We propose \textbf{OnFly}, a fully onboard zero-shot real-time AVLN system with a shared-perception dual-agent architecture, which decouples conflicting high- and low-frequency decision demands and stabilizes overall decision-making.
    \item We design a hybrid keyframe–recent-frame memory for long-term progress monitoring, which preserves global trajectory context and ensures stable KV-cache usage for reliable long-horizon observation.
    \item We present a semantic–geometric verifier integrated with a receding-horizon local planner to refine safe targets and generate optimized collision-free trajectories, balancing navigation efficiency and flight safety.
    \item We conduct extensive evaluations in simulation and real-world onboard flight experiments, with all modules running onboard the UAV, and will release our code to support reproducible AVLN research.
\end{enumerate}

\vspace{-0.1cm}

\section{Related Work}
\label{sec:related_work}
\subsection{Supervised Aerial Vision-Language Navigation}
Supervised AVLN has recently attracted growing attention. Most existing approaches follow training paradigms from ground-based VLN and rely on task-specific supervision (e.g., annotated trajectories or waypoint/action labels) to map “instruction–observation” inputs into executable decisions such as waypoints, trajectories, or actions \cite{liu2023aerialvln,gao2025openfly,wang2025uav,sautenkov2025uav,wang2024towards}. Broadly, these methods can be grouped into several representative directions: one line focuses on dataset construction and supervision design to obtain more reliable training signals and improve navigation performance \cite{liu2023aerialvln,lee2024citynav}; another line introduces intermediate representations such as waypoints, structuring the decision process as “waypoint prediction–trajectory completion/post-processing” to enhance trajectory coherence and execution stability \cite{gao2025openfly,wang2025uav,sautenkov2025uav}. Meanwhile, for complex outdoor environments, some studies further incorporate geometric cues such as depth to improve obstacle avoidance and environmental adaptability \cite{sun2026autofly}. While these methods perform well on benchmarks, limited task-data scale and scene coverage can reduce robustness under distribution shifts in open environments. Accordingly, researchers have begun to explore zero-shot AVLN with less reliance on task-specific supervision or fine-tuning.

\subsection{Zero-Shot Aerial Vision-Language Navigation}

With the continued advancement of VLMs and large language models (LLMs), zero-shot AVLN has attracted increasing attention in recent years \cite{hu2025see,lykov2025cognitivedrone,nasiriany2024pivot}. Early zero-shot AVLN paradigms mainly relied on LLMs to parse navigation instructions and decide flight directions, and combined detectors \cite{liu2024grounding} for target selection \cite{chen2023typefly}; however, their performance is overly dependent on the generalization capability of detectors, which tends to fail when the detectors cannot adequately cover the target scenarios. Another subset of works drew on the waypoint prediction idea of ground-based VLN, compressing the solution space by enumerating waypoints on images and guiding VLMs to select waypoints on images to achieve flight \cite{zhang2025logisticsvln,nasiriany2024pivot}. Such methods are overly reliant on the waypoint generation module, and inappropriate generation of waypoint candidates will constrain the model’s decision-making to an erroneous space, thereby limiting their ultimate performance. Recently, See, Point, Fly (SPF) \cite{hu2025see} reformulates the AVLN task as grounding, where VLMs directly output 2D waypoints on the image and then lift them to 3D for execution.

Although the above methods have advanced zero-shot AVLN, several common issues remain in real-world deployment. First, many systems rely on cloud-based inference, making navigation sensitive to network fluctuations and response latency; this often disrupts control continuity and causes noticeable pauses \cite{sun2026autofly}. Second, existing methods frequently lack explicit progress monitoring: while the system continuously outputs actions, it is difficult to timely determine whether the goal has been reached, whether it is drifting away from the instruction, or when to terminate or recover, which can lead to redundant actions or delayed termination. Finally, safety and efficiency are still hard to reconcile: one line of methods prioritizes safety via high-frequency short-horizon iterative decisions, which tends to induce stop-and-go behavior and reduce efficiency \cite{nasiriany2024pivot,chen2023typefly}; another line prioritizes efficiency by predicting longer-range targets, yet under noise and environmental uncertainty these targets are more likely to fall into high-risk regions, making safety difficult to guarantee \cite{hu2025see}.

\section{Problem Formulation}
\label{sec:problem_formulation}
Given a natural-language instruction and online visual observations, the system should follow the instruction to reach the final goal safely and efficiently, and autonomously stop upon arrival. Formally, given a natural-language instruction $\mathcal{L}$, a first-person RGB-D camera mounted on the agent, and an odometer sensor that provides its displacement and attitude relative to the starting point, the agent receives multimodal observations $O_t=\{\mathrm{rgb}_t,\,\mathrm{depth}_t,\,\mathrm{pose}_t\}$. The goal is to enable the system to reliably and efficiently execute the flight task specified by $\mathcal{L}$ fully onboard, autonomously determine task completion and trigger a termination signal upon reaching the final target, while strictly satisfying safety constraints (e.g., collision-free) and maintaining high efficiency throughout the flight.

\vspace{-0.1cm}

\section{Methodology}
\label{sec:method}

Fig.~\ref{fig:system} overviews our zero-shot onboard AVLN system. Given a long-horizon instruction $\mathcal{L}$, \emph{Task Management} decomposes it into a subtask queue and outputs the current subtask $\mathcal{L}_i$ (Sec.~\ref{sec:tasks_manager}). Conditioned on $O_t$ and $\mathcal{L}_i$, a shared-perception \emph{Dual-Agent} design separates high-rate goal proposing from low-rate progress monitoring (Sec.~\ref{sec:dual_agent}): the \emph{Decision Agent} predicts an image-space target point and proposes candidate goals with a lightweight history-point cue (Sec.~\ref{sec:dec_agent}), while the \emph{Monitoring Agent} tracks progress with a \textit{Hybrid Memory} that preserves long-horizon trajectory context and supports prefix-stable KV-cache reuse, and gates execution via task-state signals (Sec.~\ref{sec:memory_agent}). Both agents share ViT features with separate KV-cache to reduce redundant computation (Sec.~\ref{sec:dual_agent_coordination}). Candidate goals are then passed to a local planning module with semantic--geometric verification, which refines instruction-consistent safe goals and generates executable trajectories (Sec.~\ref{sec:verifier}).

\subsection{Dual-Agent System}
\label{sec:dual_agent}

Previous methods often couple high-frequency control decisions and low-frequency state/progress assessment within a single inference stream. Due to the mismatch in update rates, low-frequency analysis can block high-frequency target generation, leading to misalignment and unstable decisions. Moreover, because both demands share the same context and compute but pursue different goals, this coupling introduces objective conflicts and resource contention \cite{kim2025towards}.
To address this, we adopt a dual-agent architecture that decouples the inference streams: a high-frequency decision agent outputs real-time flight targets, while a low-frequency monitoring agent tracks task progress and state.

\subsubsection{Decision Agent.}
\label{sec:dec_agent}
We formulate decision-making as goal-point prediction: the VLM outputs a target point in the image coordinate system, which is then mapped to a 3D navigation goal used by the planning/control module. Compared to directly predicting discrete actions, this representation more naturally supports smooth and continuous motion~\cite{hu2025see}. 

To maintain high-frequency inference efficiency and mitigate the temporal ``amnesia'' of single-frame decision-making, we use a structured decision prompt \(\mathcal{P}_{\text{dec}}\) for the \textit{Decision Agent} (Fig.~\ref{fig:system}). The prompt defines the agent role, task context, and a strict output format, requiring the next waypoint to be returned as an integer \((x,y)\) in image coordinates under basic execution constraints. We further augment \(\mathcal{P}_{\text{dec}}\) with a lightweight history cue by reprojecting the previous 3D goal onto the current image and appending its 2D pixel coordinate as a \emph{history point}, which incurs negligible overhead. The next waypoint is then predicted by the VLM conditioned on \(\mathcal{P}_{\text{dec}}\), the current subtask \(\mathcal{L}_i\), and the observation \(O_t\), i.e., \((p_x,p_y) = \mathrm{VLM}\!\left(\mathcal{P}_{\text{dec}},\, \mathcal{L}_i,\, O_t\right)\).

We package the candidate goal as
$G_c=\big[p_x,\,p_y,\allowbreak \mathrm{depth}_t,\allowbreak \mathbf{f}_t,\allowbreak \mathrm{pose}_t\big]$,
where $\mathrm{depth}_t$ is the depth map at time $t$, $\mathbf{f}_t$ denotes the cached ViT feature associated with the candidate (Sec.~\ref{sec:dual_agent_coordination}), and $\mathrm{pose}_t$ is the current pose from odometry.
$G_c$ is then passed to the downstream semantic--geometric verification module for further validation.

\vspace{-0.1cm}

\subsubsection{Monitoring Agent with Hybrid Memory.}
\label{sec:memory_agent}
The \textit{Monitoring Agent} tracks the UAV's task-stage progress and outputs task-status signals given the current sub-instruction $\mathcal{L}_i$ and a compact monitoring memory $\mathcal{M}_t$. We condition the VLM on a monitoring prompt $\mathcal{P}_{\text{mon}}$. The prompt specifies the agent role and subtask context, enumerates completion and failure criteria (e.g., goal reached / target lost), and constrains the output to a forced-choice selection over a fixed status set. The overall framework is illustrated in Fig.~\ref{fig:system}. Formally, we model monitoring as a low-frequency state classifier, where \(s_t = \mathrm{VLM}\!\left(\mathcal{P}_{\text{mon}},\, \mathcal{L}_i,\, \mathcal{M}_t\right)\) and \(s_t \in \{\texttt{CONTINUE},\, \texttt{STOP},\, \texttt{LOST}\}\).
Here, \texttt{CONTINUE} maintains normal execution of the current subtask, \texttt{STOP} marks completion of the current subtask and terminates the episode if it is the last subtask, and \texttt{LOST} indicates that the agent is likely off-track or has lost the target, in which case the UAV first stops and then reorients to the heading of the last normal point for recovery.

To make monitoring stable for long-horizon tasks while remaining efficient, $\mathcal{M}_t$ should both preserve global context and enable effective decoder KV-cache reuse \cite{shi2024keep}. We therefore design a \textit{Hybrid Memory} scheme to construct $\mathcal{M}_t$.

\begin{figure}[t]
    \vspace{0.2cm}
        \begin{center}
        \includegraphics[width=0.90\columnwidth]{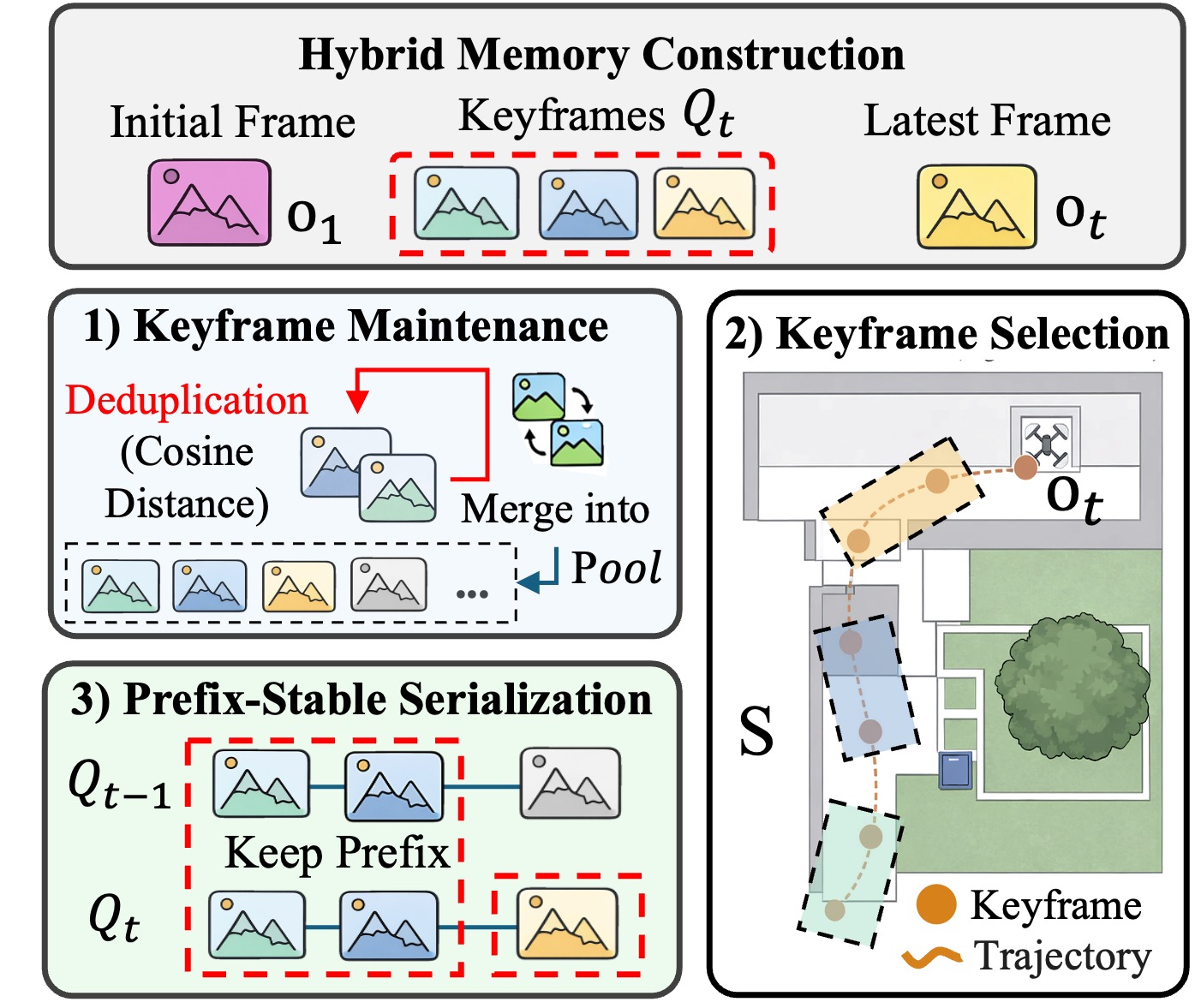}     
        \vspace{-0.6cm}
	   \end{center}
        \caption{\label{fig:keyframe}\textbf{Hybrid Memory construction.} The monitoring memory is built from the initial frame, selected keyframes, and the latest frame. It maintains a de-duplicated keyframe pool, selects representative keyframes for global coverage, and serializes them in a prefix-stable order for KV-cache reuse.}
   \vspace{-0.7cm}
\end{figure}

\textbf{Hybrid Memory.} Prior work often uses a sliding-window memory that keeps the latest $N$ observations. While simple, it gradually loses global context and reduces KV-cache reuse. KV-cache reuse benefits from keeping the leading part of the input (the \emph{input prefix}, e.g., prompt + early memory slots) identical across steps; however, a sliding window overwrites early memory slots, destabilizing the input prefix (lower cache hit rates and higher latency) and potentially reducing success when new observations overwrite long-horizon cues.
To address this, we adopt a ``first-frame + keyframe + latest-frame'' \textit{Hybrid Memory}. Let $o_t\triangleq(\mathrm{rgb}_t,\mathrm{depth}_t)$ denote the visual observation. We maintain a keyframe candidate pool $\mathcal{U}_t$ and construct an ordered keyframe list $\mathcal{Q}_t$ with prefix-stable prompting. We then form the \textit{Hybrid Memory} as $\mathcal{M}_t=[\,o_1,\mathcal{Q}_t,o_t\,]$, where $o_1$ is the initial frame and $o_t$ is the latest frame (see Fig.~\ref{fig:keyframe} for an overview). The construction consists of three steps: keyframe maintenance, keyframe selection, and prefix-stable serialization.

\emph{1) \textbf{Keyframe maintenance.}} At each step $t$, we generate a set of new keyframe candidates $\mathcal{C}_t$ based on accumulated translation/rotation thresholds, and remove near-duplicates within the set under a threshold $\epsilon$ via $\mathcal{C}^{\mathrm{new}}_t=\mathrm{Dedup}(\mathcal{C}_t,\epsilon)$, where similarity is computed as cosine distance on cached ViT features. We then merge $\mathcal{C}^{\mathrm{new}}_t$ into the pool with de-duplication against $\mathcal{U}_{t-1}$: for each $k\in\mathcal{C}^{\mathrm{new}}_t$, we first restrict the search to the $L$ nearest pool items by geometric distance and then find the most similar entry $u\in\mathcal{U}_{t-1}$ under the same feature similarity metric; if the distance is below $\epsilon$, we treat it as a duplicate and keep the existing pool entry unchanged; otherwise, we append $k$ as a new pool item. Formally, $\mathcal{U}_t=\mathrm{Merge}(\mathcal{U}_{t-1},\mathcal{C}^{\mathrm{new}}_t,\epsilon)$.

\emph{2) \textbf{Keyframe selection.}} To keep keyframes stable over time while maintaining global coverage, we partition the traversed distance $[0,D_t]$ into $S$ equal-length segments ($S$ is fixed) and assign each pool item in $\mathcal{U}_t$ to a segment by its odometry distance. We select a \emph{segment winner} for each segment using a sticky rule: we reuse the previous winner if it remains in the same segment; otherwise, we re-select the candidate closest to the segment center, with tie-breaking that favors recently updated entries. If a segment is empty, we fill the missing slot during serialization to keep $|\mathcal{Q}_t|=S$ fixed.

\emph{3) \textbf{Prefix-stable serialization.}} To enable KV-cache reuse, we keep the leading part of the input as stable as possible across steps. Given the per-segment winners, we construct an ordered keyframe list $\mathcal{Q}_t$ (length $S$) by (i) preserving the longest valid prefix of $\mathcal{Q}_{t-1}$, (ii) appending the current winners with de-duplication, and (iii) filling remaining slots with unused recent pool items (e.g., when some segments are empty or winners are duplicated). We then sort $\mathcal{Q}_t$ chronologically before input, yielding $\mathcal{M}_t=[\,o_1,\mathcal{Q}_t,o_t\,]$.

\vspace{-0.1cm}

\subsubsection{Dual-Agent Coordination.}
\label{sec:dual_agent_coordination}
The two agents share a single perception backbone and reuse cached ViT features, avoiding duplicate visual encoding. Their inference contexts are decoupled: each agent uses its own prompt/memory and maintains a separate KV-cache to match its input structure and update rate. The \textit{Monitoring Agent} runs asynchronously at a lower frequency and outputs a signal $s_t$ that modulates execution. The \textit{Decision Agent} runs at high frequency to produce real-time goal points from the observation.

\begin{figure*}[t]
    \vspace{0.2cm}
        \begin{center}
        \includegraphics[width=0.99\textwidth]{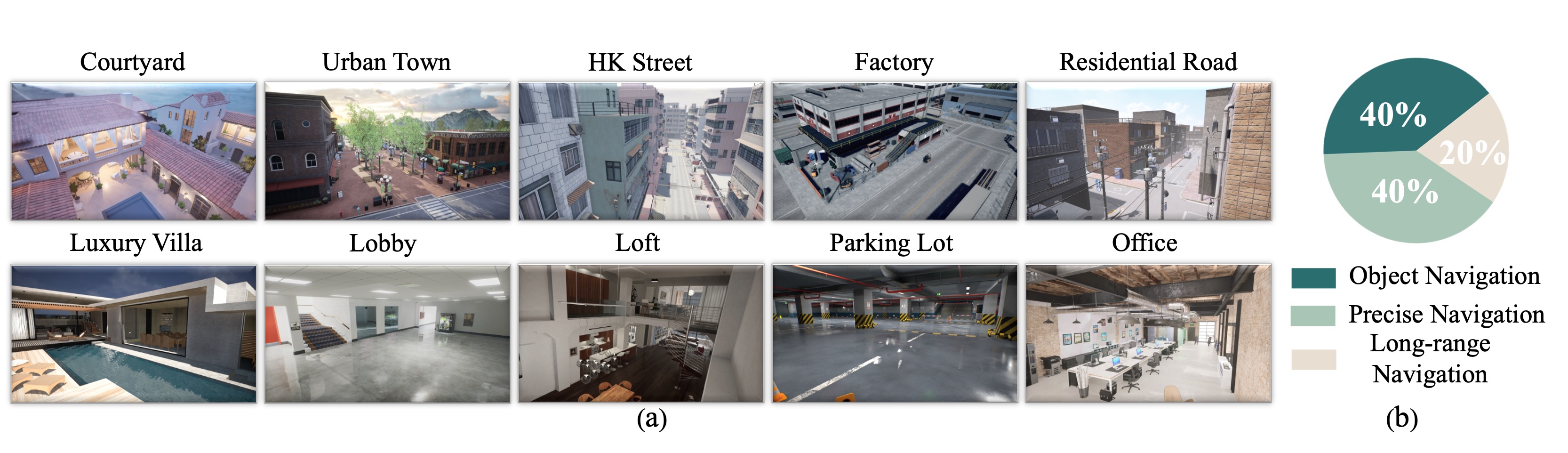}     
        \vspace{-0.6cm}
	   \end{center}
        \caption{\label{fig:sim_env}\textbf{Simulation environments for benchmarking.} (a) Examples of the 10 high-fidelity Unreal Engine scenes, covering diverse indoor and outdoor settings. (b) Task composition of the benchmark, including object navigation, precise navigation, and long-range navigation.}
    \vspace{-0.6cm}
\end{figure*}

\subsection{Semantic-Geometric Verified Local Planning}
\label{sec:verifier}

Since the VLM-predicted target primarily conveys semantic intent and does not guarantee geometric feasibility, it may fall on obstacles or other non-traversable regions. Moreover, a purely geometric correction to enforce traversability can still violate the instruction’s semantics. To address this, we introduce a semantic--geometric verifier that refines the 2D target into a traversable, instruction-consistent goal. A receding-horizon local planner then optimizes a dynamically feasible, collision-free trajectory toward this goal.

\textbf{Semantic-Geometric Verified Local Planning.}
We verify and refine the VLM-predicted target in two steps: (1) \emph{semantic-guided target refinement} to correct local misplacement while staying within a semantically consistent region, and (2) \emph{bearing-aware geometric gating} to avoid aggressive forward motion when the target lies near the image boundary.

\textbf{1) \textit{Semantic-guided target refinement.}}
Given $G_c=\big[p_x,\,p_y,\,\mathrm{depth}_t,\,\mathbf{f}_t,\,\mathrm{pose}_t\big]$, we refine the image target $\mathbf{p}=(p_x,p_y)$ to a semantically consistent and depth-feasible pixel $\mathbf{p}'$.
Let $d=\mathrm{depth}_t(\mathbf{p})$. If $d>D_{\max}$, we skip local refinement since depth and feature similarity are less reliable at long range (i.e., $\mathbf{p}'=\mathbf{p}$); otherwise, we reuse $\mathbf{f}_t$ and compute a similarity mask $\mathcal{M}_{\mathbf{p}}$ within a local ROI around $\mathbf{p}$:
\begin{equation}
\mathcal{M}_{\mathbf{p}}
=\left\{\mathbf{q}\in \mathrm{ROI}(\mathbf{p}) \ \middle|\ \mathrm{sim}\!\left(\mathbf{f}_t(\mathbf{q}),\mathbf{f}_t(\mathbf{p})\right)\ge \tau \right\}.
\end{equation}
Since $\mathcal{M}_{\mathbf{p}}$ may contain multiple disjoint clusters within the ROI, we treat it as a binary mask and partition it into connected regions; we keep only the region that contains the original target pixel $\mathbf{p}$:
$\mathcal{C}_{\mathbf{p}}=\mathrm{CC}\!\left(\mathcal{M}_{\mathbf{p}},\mathbf{p}\right)$.
We then search within $\mathcal{C}_{\mathbf{p}}$ for the nearest pixel that passes a depth-clearance check.
Let $\mathcal{S}_t$ denote the set of depth-feasible pixels at time $t$.
We denote the selected pixel as $\mathbf{p}'=\arg\min_{\mathbf{q}\in\mathcal{C}_{\mathbf{p}}\cap \mathcal{S}_t}\|\mathbf{q}-\mathbf{p}\|_2$.
We implement $\mathcal{S}_t$ by thresholding $\mathrm{depth}_t$ to obtain an obstacle mask and then dilating it with radius $r$ to enforce clearance; pixels outside the dilated obstacle mask are marked feasible.
We denote the effective depth at the refined target as $d'=\min(\mathrm{depth}_t(\mathbf{p}'),D_{\max})$.

\textbf{2) \textit{Bearing-aware geometric gating.}}
When the predicted target falls near the image boundary, it typically corresponds to a large viewing bearing, suggesting that the robot should first reduce the bearing (i.e., turn) rather than keep moving forward. If we directly use its depth as a long-range 3D navigation goal, the planner may produce excessive forward progress while the yaw is still misaligned with the target bearing, causing the trajectory to deviate from the intended path.
Let $\mathbf{u}'=[p_x',\,p_y',\,1]^\top$ be the refined pixel in homogeneous form, and define the bearing as $\theta=\arctan\!\left(\frac{p_x'-c_x}{f_x}\right)$, where $f_x$ and $c_x$ are the horizontal focal length and principal point of $K$. Let $\theta_{\max}$ denote the half horizontal field-of-view (e.g., $\theta_{\max}=\mathrm{FOV}_x/2$). We then compute the gated forward range as
$d_f=d'\exp\!\left(-\frac{1}{2}\left(\frac{\theta}{\sigma_\theta\theta_{\max}}\right)^2\right)$,
where $\sigma_\theta$ controls how quickly the range decays with increasing $|\theta|$. The corresponding 3D target in the camera frame is $\mathbf{X}_{\text{camera}}=d_f K^{-1}\mathbf{u}'$. Finally, we transform it to the odometry frame using the odometry pose $T$ to obtain the navigation goal $G_p$ for the downstream local planner.

\vspace{-0.1cm}

\subsubsection{Local Planner.}
After obtaining the 3D navigation goal \(G_p\), we adopt Fast Planner~\cite{zhou2019robust} to generate a short-horizon local path. The system maintains a local ESDF map for obstacle avoidance and enforces UAV dynamic constraints to ensure trajectory executability. In addition, we regularize yaw to align with the goal bearing during obstacle avoidance (instead of the tangential heading). Specifically, let \(\psi_t\) denote the UAV yaw, let \((x_t,y_t)\) denote the current UAV position in the odometry frame, and let \(\psi_g=\mathrm{atan2}(G_{p,y}-y_t,\,G_{p,x}-x_t)\) denote the goal bearing in the odometry frame. We add a yaw penalty \(J_\psi = \lambda_\psi \, \big\|\mathrm{wrap}(\psi_t-\psi_g)\big\|^2\), where \(\mathrm{wrap}(\cdot)\) maps angles to \((-\pi,\pi]\).

\vspace{-0.15cm}

\subsection{Task Management}
\label{sec:tasks_manager}
We maintain a Task Manager that manages a queue of subtasks. Given an instruction \(\mathcal{L}\), it decomposes the instruction into a sequence of subtasks \(\{\mathcal{L}_0, \mathcal{L}_1, \ldots, \mathcal{L}_n\}\) using the onboard VLM as a one-time computation upon instruction arrival. Once the Task Manager receives a stable \texttt{STOP} signal from the Monitoring Agent for the current subtask, it advances to the next subtask \(\mathcal{L}_{i+1}\) and repeats this process until all subtasks are completed.
\vspace{-0.15cm}

\subsection{Deployment}
\label{sec:deploy}
We deploy the VLM on Jetson Orin NX and accelerate onboard inference with TensorRT~\cite{tensorrt_github}.
We further apply AWQ quantization to the LLM for efficiency, while keeping the ViT in FP16 to preserve visual accuracy, cache visual features to avoid rerunning the vision encoder when the same image is referenced repeatedly, reuse the LLM's KV-cache to reduce redundant attention computation during long-context generation, and use CUDA Graphs to reduce kernel-launch overhead during decoding.
Together, these optimizations minimize workload and enable efficient multi-image, long-context inference on resource-constrained edge hardware.
\vspace{-0.15cm}

\begin{figure*}[t]
    \vspace{0.2cm}
        \begin{center}
        \includegraphics[width=0.99\textwidth]{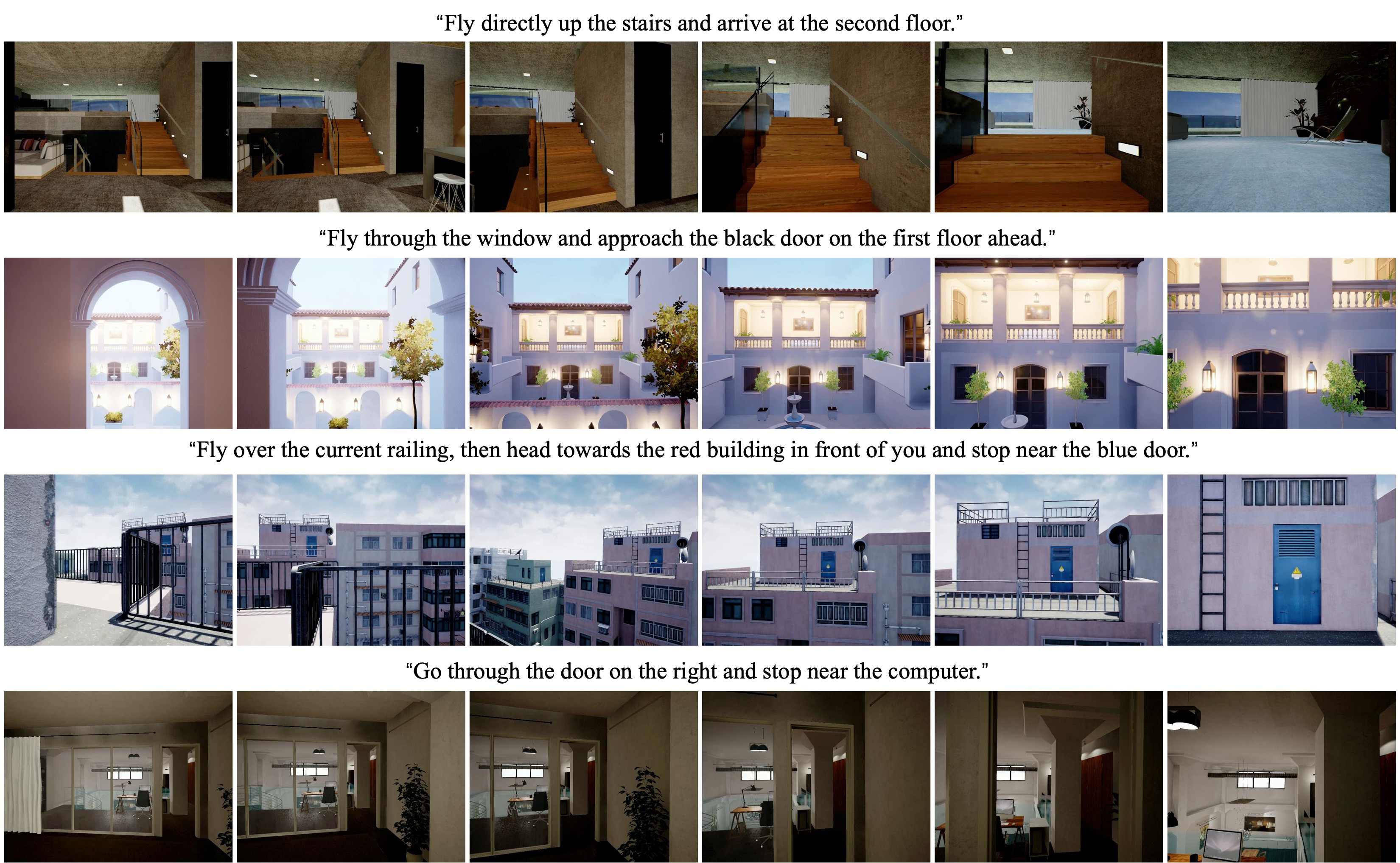}     
        \vspace{-0.5cm}
	   \end{center}
       \caption{\label{fig:sim_exp} \textbf{Simulation visualizations.} Each row shows a navigation episode, with the language instruction at the top and representative first-person UAV keyframes along the trajectory.}
   \vspace{-0.6cm}
\end{figure*}

\section{Experiments}
\label{sec:exp}

\subsection{Simulation Experimental Setup}
\label{sub:Implementation Details}

\textbf{Simulation Environments.} We constructed 10 high-fidelity 3D scenes in UE~4.27~\cite{epic_unreal_engine_2025} (e.g., urban blocks, factories, villas, and duplex buildings) to evaluate AVLN in complex environments. The benchmark contains 150 tasks spanning object, precise, and long-range navigation (40\%, 40\%, and 20\%). Object navigation requires reaching a specified target, precise navigation requires semantically grounded fine positioning (e.g., the pool edge), and long-range navigation spans multiple areas. Real trajectories were collected from the first-person view of professional pilots. Task descriptions were generated from paired images using Qwen3-VL-235B-A22B~\cite{bai2025qwen3} and manually verified. Environment examples and task composition are shown in Fig.~\ref{fig:sim_env}.

\textbf{Metrics.} Following prior work~\cite{hu2025see,sun2026autofly}, we adopt four commonly used metrics: Success Rate (SR), Oracle Success Rate (OSR), Collision Rate (CR), and Flight Time (FT). 
We terminate an episode once a collision occurs.
\begin{itemize}
    \item \textbf{SR}: Success if the UAV stops within a predefined threshold distance to the goal region \emph{without collision}.
    \item \textbf{OSR}: Success if the UAV reaches goal region at least once along its flight trajectory.
    \item \textbf{CR}: Whether the UAV collides during the task.
    \item \textbf{FT}: Total flight time from takeoff to task completion.
\end{itemize}

\textbf{Baselines.} We compare our system with three representative zero-shot AVLN methods: \textit{TypeFly}\cite{chen2023typefly}, which leverages an LLM to generate high-level plans, obtains perception cues via an object detector, and invokes UAV skills for language-guided flight; \textit{PIVOT}\cite{nasiriany2024pivot}, which performs iterative visual prompting by overlaying candidate actions and querying a VLM to select the best option in a closed-loop manner; and \textit{See, Point, Fly (SPF)}\cite{hu2025see}, a training-free VLM framework that formulates action prediction as 2D waypoint grounding on the image and converts the grounded waypoints into UAV motion commands.

\vspace{-0.1cm}

\textbf{Implementation Details.} We use a simulated UAV in AirSim equipped with a forward-facing camera (FOV: 90°$\times$60°, 720p). The UAV dynamics are constrained by a maximum velocity $v_m=0.6$ m/s, maximum acceleration $a_m=0.6$ m/s$^2$, and maximum yaw rate $\omega_m=0.4$ rad/s. Following common practice, we set the maximum depth to \(D_{\max}=7\,\mathrm{m}\). The Monitoring Agent runs once every 2~s. For \textit{Hybrid Memory}, we set the keyframe budget to $S=4$. In semantic--geometric verification, we use a similarity threshold $\tau=0.5$ and dilate the obstacle mask with radius $r=0.2$ to enforce clearance. For a fair comparison, we use the same VLM, Qwen3-VL-4B-AWQ~\cite{bai2025qwen3}, for inference across all methods. For \textit{TypeFly}, we adopt YOLOE~\cite{wang2025yoloe} as the object detector to provide perception cues. To support a unified deployment and evaluation protocol, we add lightweight stopping interfaces to baselines without native stop signals (e.g., a stop skill for TypeFly and zero-depth output for SPF). These adaptations only standardize episode termination and do not alter the baselines’ core decision logic. All experiments are conducted on the same machine with an Intel Core i9-13900K, an NVIDIA GeForce RTX 4090 GPU, and 64~GB RAM. Following prior work, the SR threshold is set to 5~m. Each episode has a 70~s time limit and is executed three times. We report the average results.

\begin{figure*}[t]
    \vspace{0.2cm}
        \begin{center}
        \includegraphics[width=0.99\textwidth]{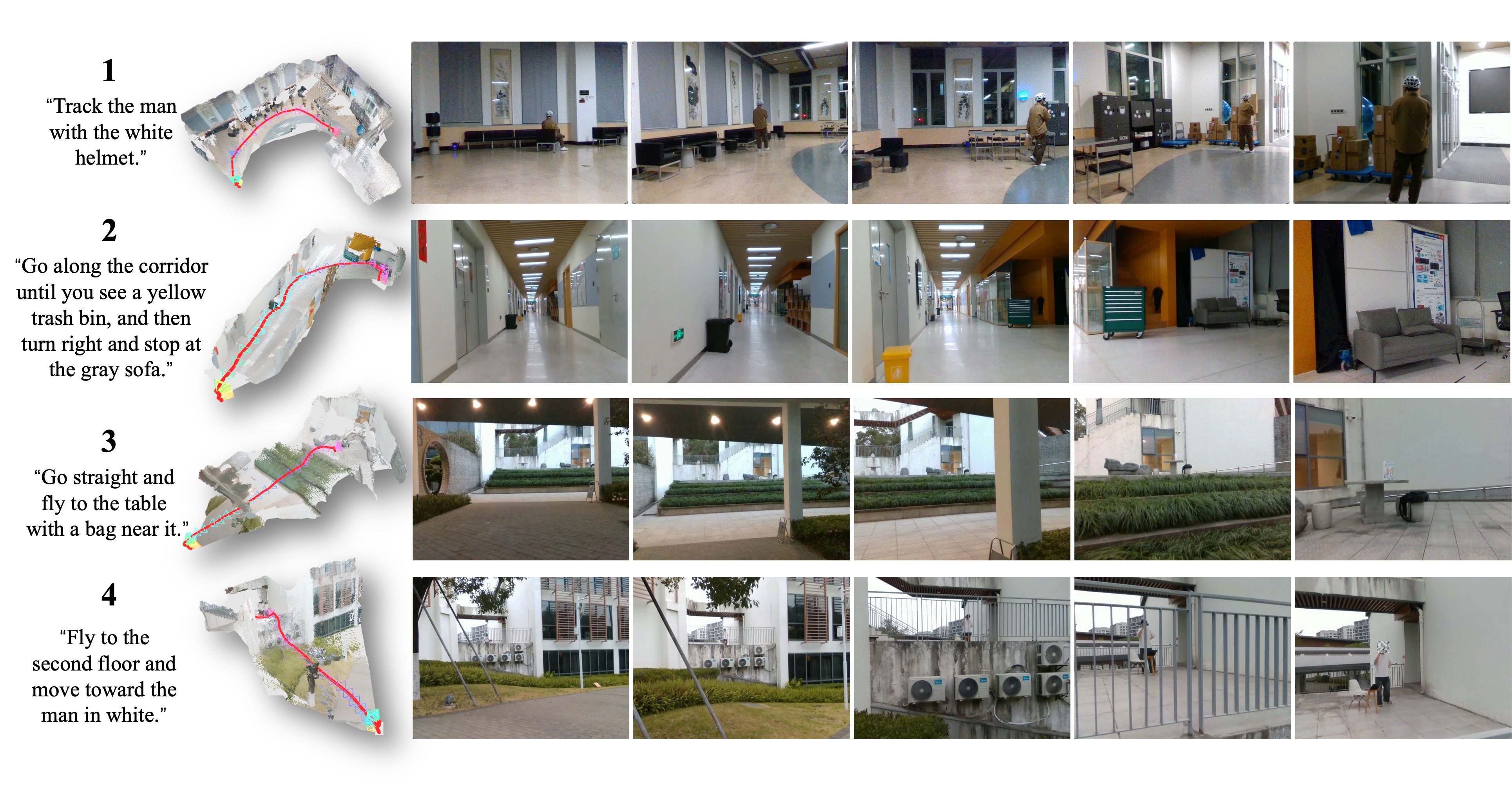}     
        \vspace{-1.0cm}
	   \end{center}
        \caption{\label{fig:realworld}\textbf{Real-world flight experiments with language instructions.} Each row corresponds to one task. The middle panel shows the executed 3D trajectory, and the right panel presents representative first-person RGB snapshots, with the corresponding language instruction shown on the left. The tasks include pedestrian tracking along a roughly circular route, long-range corridor navigation with sparse landmarks, precise outdoor navigation across a stepped lawn edge, and multi-level navigation with obstacle avoidance to reach a second-floor platform and stop near the specified person. Additional results are provided in the supplementary video.}
   \vspace{-0.5cm}
\end{figure*}

\begin{table}[t]
\vspace{0.2cm}
\renewcommand\arraystretch{1.2}
\tabcolsep=2.0mm
\caption{Zero-shot AVLN Benchmark Results.}
\vspace{-0.2cm}
\centering
\resizebox{\columnwidth}{!}{%
\begin{tabular}{c|cccc}
\hline
\textbf{Method}
& \textbf{SR} (\%) $\uparrow$
& \textbf{OSR} (\%) $\uparrow$
& \textbf{CR} (\%) $\downarrow$
& \textbf{FT} (s) $\downarrow$ \\
\hline \hline
TypeFly\cite{chen2023typefly}  & 5.7 & 17.5 & 66.5  & 35.4 \\
PIVOT \cite{nasiriany2024pivot}                         & 10.3 & 44.2 & 64.1  & 32.9\\
SPF\cite{hu2025see}  & 26.4 & 61.5 & 42.7  & 39.2 \\
\rowcolor{gray!15} Ours & \textbf{67.8} & \textbf{78.1} & \textbf{2.7} & \textbf{27.1} \\
\hline
\end{tabular}%
}
\label{table:bmk_all}
\vspace{-0.3cm}
\end{table}

\begin{table}[t]
\vspace{0.2cm}
\renewcommand\arraystretch{1.2}
\tabcolsep=2.0mm
\caption{Obstacle-Avoidance Planner Control Study.}
\vspace{-0.2cm}
\centering
\resizebox{\columnwidth}{!}{%
\begin{tabular}{c|cccc}
\hline
\textbf{Method}
& \textbf{SR} (\%) $\uparrow$
& \textbf{OSR} (\%) $\uparrow$
& \textbf{CR} (\%) $\downarrow$
& \textbf{FT} (s) $\downarrow$ \\
\hline \hline
SPF w/ planner  & 43.2 & \textbf{80.3} & 4.7  & 50.2 \\
\rowcolor{gray!15} Ours                          & \textbf{67.8} & 78.1 & \textbf{2.7} & \textbf{27.1} \\
\hline
\end{tabular}%
}
\label{table:bmk_planner}
\vspace{-0.5cm}
\end{table}

\subsection{Comparison with State-of-the-art}
\label{sub:bmk}

We compare our method with three SOTA zero-shot AVLN baselines (TypeFly, PIVOT, and SPF) under the same simulation setup. As shown in Table~\ref{table:bmk_all}, our system achieves a better balance of success, safety, and efficiency in terms of SR, OSR, CR, and FT. Qualitative results of our method in simulation are shown in Fig.~\ref{fig:sim_exp}. Specifically, Table~\ref{table:bmk_all} shows that the selected baselines exhibit markedly different performance from our proposed system. \textit{TypeFly} relies on detector-provided visual cues and an LLM to invoke UAV skills; however, its performance is brittle to missed detections, which can quickly degrade planning and execution, leading to low success and a high collision rate. \textit{PIVOT} iteratively overlays candidate actions and asks a VLM to choose among them, but this indirect use of the VLM limits its ability to fully exploit instruction understanding and visual reasoning, resulting in limited success, frequent collisions, and reduced efficiency. By contrast, our \textit{Decision Agent} directly predicts navigation goals, maximizing the utility of the VLM while stabilizing decision-making. \textit{SPF} performs relatively better overall, yet it still suffers from collisions and prolonged execution (e.g., the longest flight time), suggesting inefficiencies and possibly delayed termination. 

\textbf{Planner-Controlled Study.} To rule out collision-induced failures, we further evaluate SPF with an obstacle-avoidance planner (``SPF w/ planner''). Table~\ref{table:bmk_planner} shows that this greatly reduces CR and improves OSR and SR. However, its SR remains substantially below ours, likely due to delayed stopping caused by imperfect progress estimation. The added detours also increase travel distance and time, which improves OSR but leads to longer FT.

Overall, our dual-agent design decouples real-time control from global monitoring, and our semantic-geometric verification and safety-aware planning enables timely stopping and safer, more efficient navigation, achieving higher SR/OSR, lower CR, and shorter FT than all baselines.


\begin{table}[t]
\vspace{0.10cm}
\renewcommand\arraystretch{1.15}
\tabcolsep=1.4mm
\caption{Ablation Study Results.}
\vspace{-0.25cm}
\centering
\resizebox{\columnwidth}{!}{%
\begin{tabular}{l|cccc}
\hline
\textbf{Setting}
& \textbf{SR} (\%) $\uparrow$
& \textbf{OSR} (\%) $\uparrow$
& \textbf{CR} (\%) $\downarrow$
& \textbf{FT} (s) $\downarrow$ \\
\hline
\multicolumn{5}{c}{\textit{Architecture Ablation}} \\
\hline
\rowcolor{gray!15}Ours              & \textbf{67.8} & \textbf{78.1} & 2.7 & \textbf{27.1} \\
w/o Dual Agent                      & 48.6          & 77.9          & 2.9 & 44.1 \\
w/o Verification                    & 63.7          & 72.6          & 3.2 & 28.3 \\
w/o Planner                         & 35.8          & 69.4          & 37.5 & 32.2 \\
\hline
\multicolumn{5}{c}{\textit{Memory Ablation}} \\
\hline
Time Sampling                       & 64.2         & 78.4 & 2.9 & 27.4 \\
Sliding Windows                     & 60.4         & \textbf{79.5}         & 2.8 & 28.3 \\
\rowcolor{gray!15}Hybrid Memory     & \textbf{67.8} & 78.1          & \textbf{2.7} & \textbf{27.1} \\
\hline
\multicolumn{5}{c}{\textit{VLM Backbone Ablation}} \\
\hline
Qwen3-VL-2B                         & 43.4          & 53.6          & 2.6 & \textbf{23.7} \\
Qwen3-VL-4B                         & 67.8          & 78.1          & 2.7 & 27.1 \\
Qwen3-VL-30B-A3B                    & 71.8          & 84.0          & \textbf{2.5} & 29.2 \\
\hline
\end{tabular}%
}
\label{table:ablation_all}
\vspace{-0.400cm}
\end{table}

\subsection{Ablation Study}
\label{sub:ablation_study}

\vspace{-0.1cm}
We conduct three ablation experiments: architecture-level module removal, memory design comparison, and VLM-backbone scaling, with results summarized in Table~\ref{table:ablation_all}.

\textbf{Architecture Ablation Study.}
We ablate OnFly by removing Dual Agent (\textit{w/o Dual Agent}), Verification (\textit{w/o Verification}), and Planner (\textit{w/o Planner}), as shown in Table~\ref{table:ablation_all}.
Removing the Dual Agent significantly reduces SR and increases FT, showing that decoupling high-rate target generation from low-rate progress monitoring is crucial for stable and efficient decision-making.
Disabling Verification decreases SR/OSR, confirming its role in correcting semantically inconsistent or geometrically infeasible VLM targets under noise and depth errors.
Removing the Planner sharply raises CR and reduces SR, showing ESDF-based receding-horizon planning is essential for collision-free execution.

\textbf{Memory Ablation Study.}
We compare time sampling, sliding windows, and hybrid memory, as shown in Table~\ref{table:ablation_all}. Hybrid memory achieves the best SR and FT with the lowest CR, indicating the best overall trade-off. Compared with sliding windows, it better preserves long-horizon context for final task completion. Although sliding windows and time sampling obtain slightly higher OSR, both more often fail to stop at the correct time, resulting in lower SR. Table~\ref{table:onboard_opt} further shows that hybrid memory reduces monitoring latency through more efficient KV-cache reuse.

\textbf{VLM Backbone Ablation Study.}
We replace the VLM backbone with Qwen3-VL-2B/4B/30B-A3B~\cite{bai2025qwen3} while keeping the rest of OnFly unchanged.
As shown in Table~\ref{table:ablation_all}, stronger backbones consistently improve SR/OSR with similar CR, but may incur higher runtime cost.
The 2B model also exhibits more premature stopping and less stable behavior.
These results demonstrate that OnFly scales effectively with model capability.

\begin{table}[t]
\vspace{0.15cm}
\renewcommand\arraystretch{1.22}
\tabcolsep=1.9mm
\caption{Onboard Latency Breakdown on Jetson Orin NX.}
\vspace{-0.2cm}
\centering
\resizebox{\columnwidth}{!}{%
\begin{tabular}{c|cc|cc}
\hline
\textbf{Strategy}
& \multicolumn{2}{c|}{\textbf{Decision}}
& \multicolumn{2}{c}{\textbf{Monitoring}} \\
\cline{2-5}
& \textbf{Cost} (s) $\downarrow$ & \textbf{Spd.} $\uparrow$
& \textbf{Cost} (s) $\downarrow$ & \textbf{Spd.} $\uparrow$ \\
\hline \hline
Baseline & 3.83 & 1.00$\times$ & 7.47 & 1.00$\times$ \\
Edge-Deploy      & 0.81 & 4.73$\times$ & 1.98 & 3.77$\times$ \\
\rowcolor{gray!15} w/ Hybrid Memory & 0.81 & 4.73$\times$ & 1.15 & 6.50$\times$ \\
\hline
\end{tabular}%
}
\label{table:onboard_opt}
\vspace{-0.70cm}
\end{table}

\subsection{Real-World Experiments}
\label{sub:real_world}

To evaluate real-world deployability, we conduct flight experiments on a custom quadrotor platform. We use the same speed constraint, parameter settings, and model configuration as in simulation. The platform integrates a Jetson Orin NX (16GB), a Mid360 LiDAR for state estimation, and a D435 camera for 640$\times$480 RGB-D input. The quadrotor is controlled by PX4, as shown in Fig.~\ref{fig:top}. We evaluate the system on four real-world tasks (Fig.~\ref{fig:realworld}): (i) pedestrian tracking along an approximately circular route; (ii) long-range corridor navigation to a specified target; (iii) outdoor precise navigation across a stepped lawn edge to a designated stopping point; and (iv) object navigation with obstacle avoidance, where the UAV flies to a second-floor platform, passes over a railing, and stops near a specified person. More results are provided in the supplementary video.
\vspace{-0.1cm}

\subsection{Onboard Efficiency}
\label{sub:efficiency}
Table~\ref{table:onboard_opt} reports per-step latency on Jetson Orin NX. Compared with a standard Transformers-based implementation~\cite{wolf2020transformers}, our deployment reduces both decision and monitoring latency, and \textit{Hybrid Memory} further speeds up the monitoring branch. These results support the feasibility of fully onboard execution.

\vspace{-0.15cm}

\section{Conclusion}
\label{sec:conclusion}

We presented OnFly, a fully onboard real-time zero-shot AVLN system for unknown 3D environments. OnFly decouples high-frequency goal generation from low-frequency task monitoring through a shared-perception dual-agent design, and uses \textit{Hybrid Memory} to support stable and efficient long-horizon progress assessment. We further introduced semantic--geometric verification and local replanning to refine VLM-predicted goals into executable and collision-safe trajectories. Experiments in simulation and fully onboard real-world flights demonstrated that OnFly improves task success, safety, and navigation efficiency over prior zero-shot AVLN baselines. These results suggest that combining decoupled decision-making, long-horizon monitoring, and safety-aware planning is a promising direction for deployable AVLN. Future work will focus on extending the system to longer-horizon, more challenging open-world tasks. Meanwhile, we will leverage advanced optimization techniques and higher-performance hardware to further improve inference speed and achieve more robust, responsive instruction following.









\bibliography{references}

\end{document}